%% file: main.tex
\documentclass{article}

\PassOptionsToPackage{numbers,sort&compress}{natbib}
\usepackage[final]{neurips_2026}
\makeatletter
\renewcommand{\@noticestring}{Workshop on Who Verifies the Agents? Toward Reliable Agent Development, NeurIPS 2026.}
\makeatother

\usepackage[utf8]{inputenc}
\usepackage[T1]{fontenc}
\usepackage{amsmath,amssymb,amsthm}
\usepackage{array}
\usepackage{booktabs}
\usepackage{graphicx}
\usepackage{microtype}
\usepackage{tabularx}
\usepackage{xcolor}
\usepackage[hidelinks]{hyperref}
\usepackage{url}

\graphicspath{{artifacts/}}
\theoremstyle{definition}
\newtheorem{observationequivalencetheorem}{Audit-View Non-Identifiability Theorem}
\newcommand{\safe}{\mathrm{safe}}
\newcommand{\harm}{\mathrm{harm}}
\newcolumntype{Y}{>{\raggedright\arraybackslash}X}
\newcolumntype{L}[1]{>{\raggedright\arraybackslash}p{#1}}
\newcommand{\resultheading}[1]{%
  \par\smallskip
  \noindent\textbf{#1}\par\nobreak
  \vspace{1pt}
}

\title{Post-Audit Mirage: Identifiability Limits of Offline Agent Verification}

\author{%
  Aman Behera\\
  IIT Roorkee
  \And
  Payal Payal\\
  IIT Roorkee
}

\begin{document}

\maketitle

\begin{abstract}
Offline tests can fail to reveal whether an agent update will remain safe after deployment.
We operationalize this post-audit mirage using matched safe and harmful lifecycles that expose identical verifier-visible evidence while an isolated scorer records opposite deployment truth.
Our theorem applies standard observational-equivalence and indistinguishability ideas to this executable audit boundary and states the resulting error-abstention tradeoff.
The Post-Audit Mirage benchmark makes that boundary executable across authorization, scheduling, and shared-capacity environments.
IdentifiedRangeMonitor uses fresh group-labelled harm outcomes to return deploy, hold, cannot determine, or unsupported.
Experiments show that monitoring can recover the missing distinction in favorable settings but may remain undecided for rare groups, small effects, or missing observations.
Verifier-agent evaluation must therefore test both calculation validity and whether the available evidence supports the deployment decision.
\end{abstract}

\section{Introduction}

An agent proposes a new authorization rule that passes every stored request and offline check.
In one deployment it remains safe; in another, an unobserved change in who reaches a sensitive route later harms a rare group.
If both deployments produced the same approval record, an offline verifier receives the same input in both worlds and cannot know which action is correct.

We call this the post-audit mirage: a fixed audit record can support more calculation without supplying missing deployment information.
Reusable holdouts address adaptive reuse of a sample \citep{dwork2015holdout,dwork2015adaptive,bassily2016stability}, while online testing and confidence sequences control repeated inference under stated probability laws \citep{howard2021cs,ramdas2023safe,tian2019onlinefwer,tian2019addis,fischer2024closure,fischer2024eclosed}.
These tools cannot identify a property when the permitted observations remain compatible with opposite answers.
This distinction matters when deployment changes later data or outcomes \citep{perdomo2020performative,brown2022stateful,liu2018delayed}.
Related work studies how to detect, characterize, and benchmark distribution shift \citep{rabanser2019failing,fang2022ood,koh2021wilds}.

The paper makes the following independently testable contributions.
\begin{enumerate}
  \item \textbf{Executable indistinguishability boundary:} we apply standard observational-equivalence and two-point indistinguishability ideas to an agent-audit interface, then operationalize the boundary with byte-identical public records and isolated protected truth.
  \item \textbf{Post-Audit Mirage benchmark:} an exact matched-world lifecycle benchmark isolates scorer truth from every offline rule across three environments.
  \item \textbf{IdentifiedRangeMonitor:} a fail-closed method uses fresh group-labelled outcomes to decide deploy or hold, remains undecided when evidence is insufficient, and returns unsupported when its collection assumptions fail.
\end{enumerate}

Existing agent benchmarks and realistic task environments measure task success, interaction, and sandboxed risks \citep{liu2024agentbench,ma2024agentboard,ruan2024toolemu,debenedetti2024agentdojo,zhou2024webarena,jimenez2024swebench,yao2025taubench,xie2024osworld}.
Runtime monitoring and worst-group evaluation provide complementary tools \citep{leucker2009rv,kang2020assertions,hashimoto2018fairness,kearns2018gerrymander,sagawa2020groupdro}.
Recent systems instead budget evidence across edits (SGM), test candidates sequentially (PACE), compose controllers (SEA), add an exogenous audit (SEAL), or benchmark evolutionary trajectories (SEA-Eval) \citep{wu2025sgm,shawn2026pace,sengupta2026sea,guo2026seal,jiang2026seaeval}.
Here we ask an orthogonal question: do the verifier's permitted observations identify the deployment property?

\section{Why the audit cannot decide}

Let $O$ be the complete public record available before deployment and let $Z$ be scorer-only deployment truth.
An offline rule $A$ maps $O$ and private randomness $U$ to deploy, hold, or abstain, where $U$ is independent of $O$ and has the same distribution in both worlds.
The following result directly applies standard identification and two-point indistinguishability arguments; the contribution is its executable lifecycle-audit operationalization, not a new theoretical principle \citep{rothenberg1971identification,yu1997assouad,manski2003partial}.

\begin{observationequivalencetheorem}
\label{thm:equivalence}
Suppose a safe world $P_{\safe}$ and a harmful world $P_{\harm}$ induce the same distribution for $O$ but require opposite non-abstaining decisions.
Then every offline rule $A(O,U)$ has the same decision distribution in both worlds.
If its abstention probability is $r$, then
\begin{equation}
\max\!\left\{P_{\harm}(A=\text{deploy}),P_{\safe}(A=\text{hold})\right\}
\geq \frac{1-r}{2}.
\label{eq:lower-bound}
\end{equation}
\end{observationequivalencetheorem}

Equal laws for $(O,U)$ imply equal laws for $A$.
If deploy, hold, and abstain have common probabilities $p,q,r$, the two wrong-action probabilities are $p$ and $q$, and $p+q=1-r$.
For example, if a rule abstains 20\% of the time, at least one matched world has error of at least 40\%.
The public record identifies a set of compatible answers, not a unique answer.
Appendix~\ref{app:proof} gives the complete proof.

\begin{figure}[h]
  \centering
  \includegraphics[width=\textwidth]{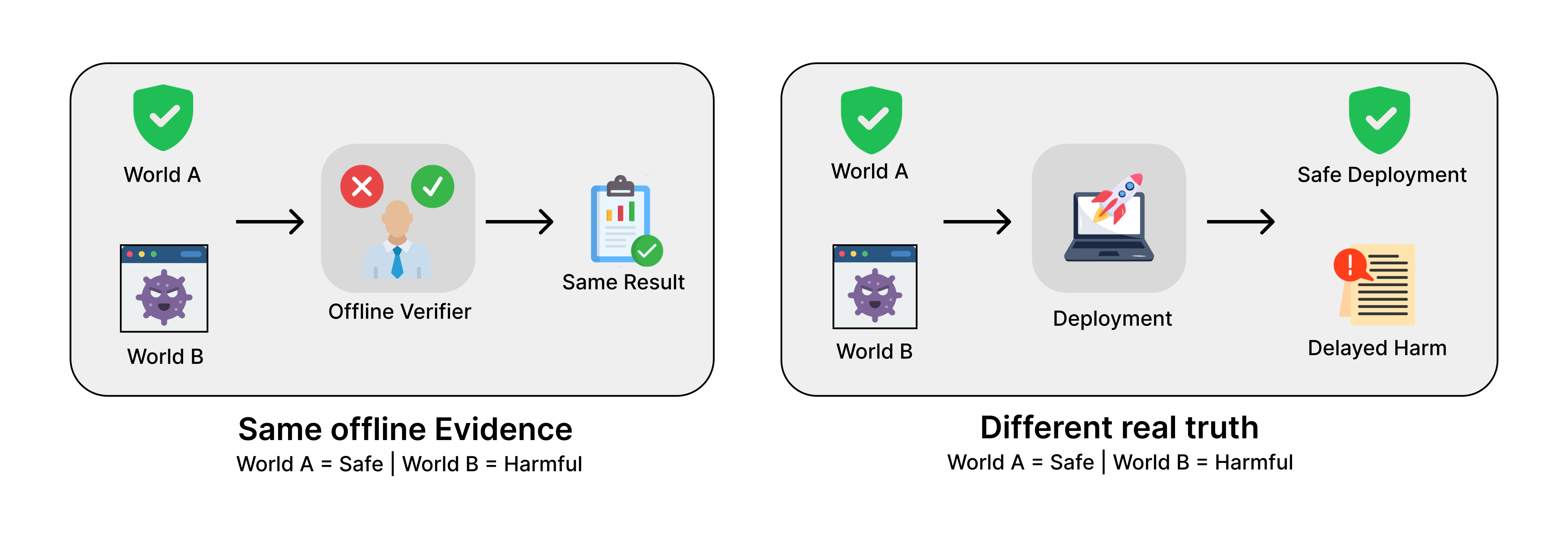}
  \caption{Matched worlds expose the same complete offline evidence, forcing the same verifier decision, while their post-update outcomes require opposite actions.
  The Safe and Harmful labels shown for World A and World B are protected scorer-only truth and are never available to the offline verifier.}
  \label{fig:architecture}
\end{figure}

\section{Benchmark and monitoring}

Each benchmark unit is executed as a safe-harmful pair for 50 update rounds.
The pair shares one serialized public lifecycle object, while the exact evaluator receives a separate realized environment object.
Construction fails unless public rounds are identical, protected answers are opposite, and every public-only method returns the same paired decision sequence.
We generate 100 AuthGate, 100 ConstraintPlan, and 72 BatchTriage pairs.

\begin{table}[t]
\caption{Exact benchmark families and their evidence boundary.
Public audit fields are method-facing; deployment effects remain scorer-only.}
\label{tab:environments}
\centering
\small
\setlength{\tabcolsep}{4pt}
\begin{tabularx}{\textwidth}{@{}L{0.14\textwidth}L{0.20\textwidth}L{0.26\textwidth}Y@{}}
\toprule
Family & Agent update & Public audit & Scorer-only deployment effect \\
\midrule
AuthGate & Change route access by group & Immediate access, audit harm, task quality & Delayed route harm, correlated incidents, live group occupancy \\
ConstraintPlan & Replace the incumbent schedule & Dependencies, deadlines, audit feasibility, objective & Cooldowns, protected blackouts, delayed violations, rare-group harm \\
BatchTriage & Change the shared-server policy & Per-group arrivals, service cost, completion & Co-arrival correlation, contention, backlog, delayed group harm \\
\bottomrule
\end{tabularx}
\end{table}

We run nine public-only diagnostic rules, ranging from always hold and fixed thresholds to adaptive spending, online e-testing, reused holdout, and transferred evidence.
These local abstractions test whether any decision changes when method-facing information is fixed; they are not faithful reproductions or rankings of complete external systems.

The monitor starts from the compatible answer set.
If both safe and harmful truth remain possible, it returns cannot determine.
After the update is fixed, it consumes binary harm outcomes for a public roster of $G$ groups.
For group $g$ and its $n$th retained outcome it spends
\begin{equation}
\alpha_{g,n}=\frac{\alpha}{G n(n+1)}, \qquad
r_{g,n}=\sqrt{\frac{\log(2/\alpha_{g,n})}{2n}}.
\label{eq:monitor}
\end{equation}
It deploys for the group-harm component only when every group upper bound is at or below the threshold, holds when any lower bound is above the threshold, and otherwise remains undecided.
The guarantee assumes independent binary harm outcomes with a stable rate within each required group, a group list fixed in advance, missingness unrelated to harm, and no change in the target outcome distribution.
An unseen required group cannot pass, and a declared biased or shifted collection design returns unsupported.
The monitor certifies only declared group harm, not feasibility, utility, undeclared groups, or complete deployment safety.
We compare the same retained counts with a pointwise Wilson score interval at each separately prespecified sample size \citep{wilson1927probable,brown2001interval} and with an anytime-valid beta-binomial mixture confidence sequence \citep{robbins1970statistical,howard2021cs}.
The Wilson comparison is fixed-sample only and is not valid for optional stopping.

\begin{figure}[t]
  \centering
  \includegraphics[width=\textwidth]{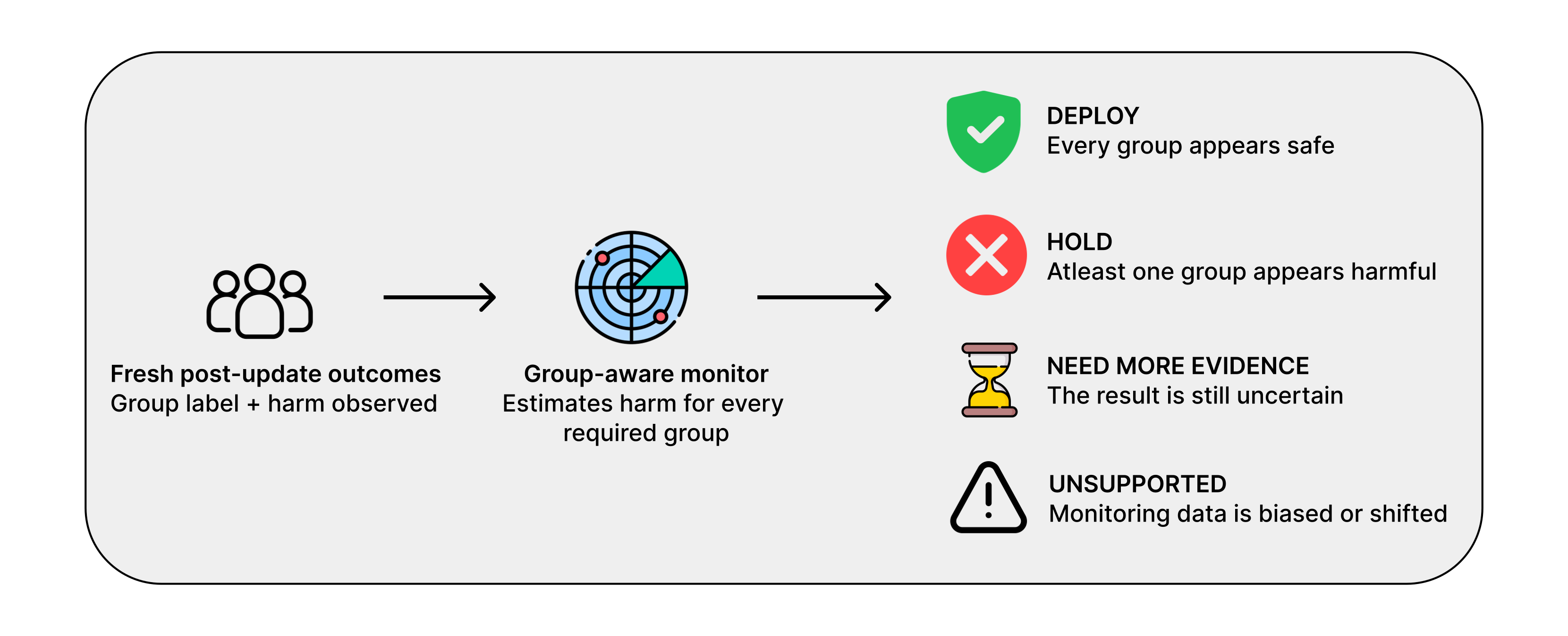}
  \caption{The fail-closed monitor uses fresh group-labelled harm outcomes after the update is fixed.
  It deploys only when every required group's upper bound is safe, holds when any lower bound is harmful, and returns \textsc{Need More Evidence}, meaning \texttt{cannot\_determine}, while the plausible harm-rate range still crosses the threshold; biased or shifted monitoring is \texttt{unsupported}.}
  \label{fig:monitor}
\end{figure}

\section{Experiments and results}

The three studies test whether identical audits force identical decisions, whether fresh outcomes restore the distinction, and when monitoring becomes informative.

\resultheading{Matched audit}
Across 272 pairs and 50 rounds per pair, all 13,600 public inputs match byte for byte, protected truth reverses, and all nine offline procedures return identical paired sequences when given the same random draws.
This verifies observation equivalence without scorer leakage; it does not show that the procedures are generally equivalent or poor.

\resultheading{Fixed-panel restoration}
For each lifecycle, we sample 20 streams per world from the full simulator's exact group frequencies and harm rates.
Each semantic lifecycle is one statistical unit; its 20 nested Monte Carlo streams are aggregated as correct only if at least 16 are correct, and are never counted as independent lifecycle observations.
At 20,000 attempted observations, all 272 safe and 272 harmful lifecycles meet the prespecified descriptive correctness rule, with 0/272 harmful lifecycles incorrectly deployed as safe in the final summary.
This is a descriptive result on a fixed synthetic panel, not a population error-rate estimate; zero observed false-safe lifecycles is not zero risk, and the monitor observes group harm rather than replaying full deployment trajectories.

\resultheading{Controlled difficulty sweep}
We cross gaps of 0.025, 0.05, and 0.10 around the 15\% threshold, rare-group prevalence of 0.02, 0.10, and 0.50, and retention, the fraction of outcomes observed, of 0.5 or 1.0, using 500 independent streams per world and cell.
Here one generated stream within a prespecified cell and world is one independent statistical unit.
With the primary Hoeffding monitor, only four settings qualify within the inspected range.
Smaller margins, rarer groups, and missing observations keep the monitor undecided longer.

\begin{figure}[!ht]
  \centering
  \includegraphics[width=\textwidth]{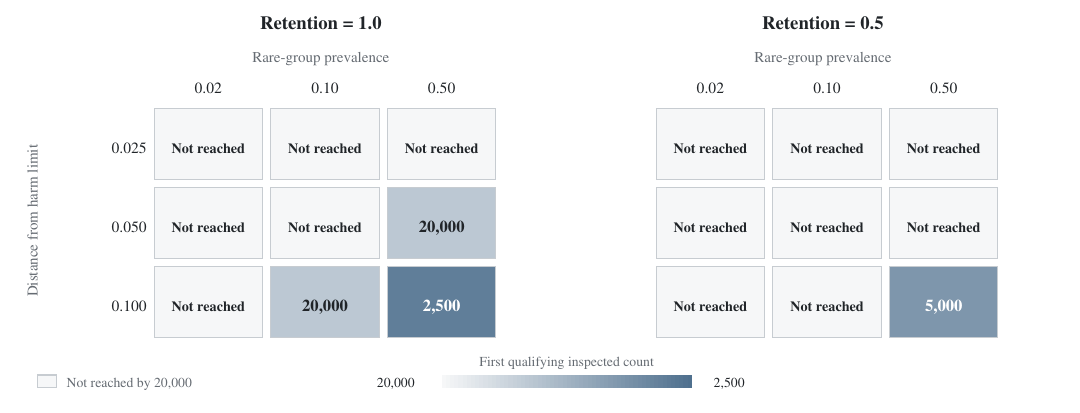}
  \caption{Fresh monitoring needed for reliable separation.
  Each cell gives the first attempted-observation count at which safe and harmful streams meet the stated correctness and harmful-approval requirements.
  ``Not reached'' means no inspected look qualifies by 20,000 attempted observations.
  Larger safety margins, more common affected groups, and more complete observations generally require less data.}
  \label{fig:controlled}
\end{figure}

A setting qualifies using one-sided exact binomial bounds computed separately at each sample size.
Released tables give all four status rates at every look and right-censored first decisions; Wilson onsets are descriptive, and no first qualifying inspected count is a formal sample-complexity estimate.
None of the three studies ranks the nine offline rules as general-purpose methods.

\resultheading{Scope}
The benchmark constructs exact observation equivalence and does not estimate its production prevalence or monitor success.
The monitor assumes independent fixed-rate binary outcomes, outcome-independent missingness, and a stable target; it cannot detect undeclared selection bias, missing-not-at-random outcomes, group omission, or target shift \citep{rubin1976missing,heckman1979selection,kearns2018gerrymander,lipton2018labelshift}.
Anytime-valid inference protects against optional stopping under its sampling assumptions but does not repair these information failures.
If a required group is unobserved, it returns \texttt{cannot\_determine}; if the declared collection design is biased or the target has shifted, it returns \texttt{unsupported}.

\resultheading{Conclusion}
The standard indistinguishability argument shows that missing deployment truth cannot be recovered from an audit, and our theorem instantiates that argument at the benchmark's audit interface.
The benchmark makes this failure executable, and the monitor tests how fresh evidence can restore a declared group-harm decision without claiming complete safety.
Verifier-agent evaluation should test both calculations and whether their evidence supports the decision.

\clearpage
\bibliographystyle{plainnat}
\bibliography{references}

\appendix

\section{Complete proof}
\label{app:proof}

Let $P_{\safe}^{O}=P_{\harm}^{O}$ and let $U$ have the same distribution in both worlds, independent of $O$.
Then the joint laws of $(O,U)$ are equal.
For every measurable decision set $B$, the mapping theorem gives
\[
P_{\safe}\{A(O,U)\in B\}=P_{\harm}\{A(O,U)\in B\}.
\]
Write the common probabilities of deploy, hold, and abstain as $p,q,r$.
Deploy is an error in the harmful world and hold is an error in the safe world, so the two error probabilities are $p$ and $q$.
Since $p+q=1-r$, $\max(p,q)\geq(p+q)/2=(1-r)/2$.
For a deterministic non-abstaining rule, the shared observation forces either deploy in both worlds or hold in both worlds, so one decision is wrong with probability one.

\section{Environment and integrity details}
\label{app:environments}

The public record contains only method-permitted fields.
The exact scorer receives a separate realized environment object with protected post-update outcomes.
The paired constructor asserts equal serialized public rounds and opposite protected answers before writing a witness.
The evaluation runner then passes only public records to the nine diagnostic procedures: always hold, greedy, fixed threshold, shrinking-budget spending, ADDIS-style spending, online closed e-testing, per-update reset, reused holdout, and transferred cross-update evidence.

Generation enforces four invariants.
Paired worlds must share all 50 public rounds, protected answers must be opposite, every public-only procedure must receive coupled inputs and randomness, and protected scorer fields must be absent from method constructors.
Failure of any invariant stops generation.

\paragraph{Operational AuthGate vignette.}
A hospital tests a rule that grants a time-limited export token to approved on-call clinicians.
The stored audit contains the same approved requests, immediate task completions, and zero immediate violations in both matched worlds.
In the safe world, the deployment gateway expires every token before the next shift.
In the harmful world, a separately configured renewal worker keeps some tokens active for three steps, and a later routing change sends more rare-group records through those sessions.
Neither configuration is present in the predeployment audit, so both worlds are operationally plausible from the verifier's permitted record even though their delayed group-harm truth is opposite.
This vignette establishes operational plausibility only and is not empirical production evidence.

\section{Monitor guarantee}
\label{app:monitor}

For each required group $g$, the monitor stores an observation count $n_g$ and harm sum $s_g$.
At every retained outcome it computes Equation~\ref{eq:monitor} and the clipped interval $[\widehat\mu_{g,n}-r_{g,n},\widehat\mu_{g,n}+r_{g,n}]$.
Within each group, retained outcomes are assumed to be independent Bernoulli draws with one fixed deployment mean.
Group arrivals and outcome-independent retention may alter counts but not the conditional outcome law.

For a fixed group and count, Hoeffding's inequality gives failure probability at most $\alpha_{g,n}/2$ on each side.
Therefore
\[
\sum_{g=1}^{G}\sum_{n=1}^{\infty}\alpha_{g,n}
=\frac{\alpha}{G}\sum_{g=1}^{G}\sum_{n=1}^{\infty}\frac{1}{n(n+1)}=\alpha.
\]
The union bound gives simultaneous coverage over all groups and observation counts under the stated model.
On this event, deploy implies that every declared group-harm mean is at most threshold, while hold implies that at least one is larger.
No conclusion follows for feasibility, utility, an undeclared group, or a shifted target.

For comparison, the fixed-sample Wilson score interval uses the standard score center and radius at each prespecified look.
It is reported pointwise and is never carried forward as an anytime-valid decision.
The beta-binomial mixture method inverts a Bernoulli mixture martingale, producing a confidence sequence that is valid uniformly over time under the same independent fixed-rate model.
All three methods receive identical group labels, observation counts, and harm counts, and all allocate the lifecycle error budget across required groups.

\section{Experimental details}
\label{app:experiments}

All generators use seed 20260821.
The matched panel has 272 semantic units and 50 rounds per unit.
Its 100 AuthGate, 100 ConstraintPlan, and 72 BatchTriage environment structures are identifier-free and distinct, but their normalized method-facing signatures number 1, 100, and 72 respectively.
AuthGate deliberately places structural variation behind one audit pattern.
The fixed panel uses 20 streams per world and unit.
The controlled panel uses 500 independent streams per world and cell.
The twelve inspected sample sizes range from 5 to 20,000 attempted observations.
The per-stream simultaneous group-and-time monitoring level is $\alpha=0.05$, and the group-harm threshold is 0.15.
The four qualifying controlled settings and their first qualifying inspected looks are listed in Section~4; the remaining 14 settings do not qualify by 20,000 attempted observations.

A controlled cell qualifies when pointwise one-sided 95\% exact lower bounds for safe and harmful correctness are at least 0.80 and the pointwise upper bound for cumulative harmful false-safe decisions is at most 0.05.
The first qualifying inspected look is a descriptive decision-onset diagnostic, not a sample-complexity estimate or a simultaneous 95\% confidence statement across looks and metrics.
The evaluation is CPU-only and uses no model training or external dataset.
One reference regeneration of the central result on an Apple M5 arm64 CPU with 10 cores and 16 GB RAM, Python 3.13.12, and eight workers took 267.30 seconds wall time, 1,894.38 seconds user CPU time, and 8.43 seconds system time.
The maximum resident set reported by \texttt{/usr/bin/time -lp} was 1.55 GB, and the command wrote 111.5 MB; it used no GPU, network, or model service.
These machine-specific measurements are not hardware requirements, and earlier development and failed exploratory runs used additional unrecorded CPU time that is excluded.

The executable release at \url{https://github.com/beingamanforever/Post-Audit-Mirage} includes the full source, independent JavaScript scorers, complete configurations, frozen synthetic records, named integrity tests, generated results, an MIT license, third-party notices, and exact reproduction commands.
From the repository root, regenerate the central result with:
\begin{center}
\begin{minipage}{0.94\textwidth}
\small\ttfamily
python -m src run-identifiability-experiments \textbackslash\\
\hspace*{1em}--config configs/identifiability.json --data-dir data \textbackslash\\
\hspace*{1em}--output-dir reproduced/data \textbackslash\\
\hspace*{1em}--artifacts-dir reproduced/artifacts --workers 8
\end{minipage}
\end{center}

\input{status_rates}

The benchmark may help deployment teams refuse unsupported decisions and expose missing groups.
Its conservative monitor can also delay beneficial updates, and a declared group roster can omit affected people.
It should inform human governance rather than autonomously authorize deployment.

\input{checklist}

\end{document}

%% file: status_rates.tex
\section{Complete status and decision-time reporting}
The fixed-panel tables aggregate the 272 semantic lifecycles per world, after applying the prespecified 16-of-20 nested-stream rule within each lifecycle.
The controlled tables aggregate the equally sized 18-cell grid, totaling 9,000 independent streams per world; they describe this grid and not a production population.
All rates are generated from \texttt{identifiability\_summary.json}.
Unsupported was 0.0\% at every reported look; declared assumption violations are reported separately as unsupported.

\begin{table}[t]
\centering
\scriptsize
\caption{Fixed-panel Hoeffding monitor status rates at every evaluated look. These 272 heterogeneous synthetic lifecycles per world form a fixed descriptive panel, not independent draws supporting a population error-rate estimate.}
\label{tab:fixed-status-rates}
\setlength{\tabcolsep}{3.4pt}
\begin{tabular}{r rrr rrr}
\toprule
& \multicolumn{3}{c}{Safe world (\%)} & \multicolumn{3}{c}{Harmful world (\%)} \\
Look & Deploy & Hold & Cannot & Deploy & Hold & Cannot \\
\midrule
5 & 0.0 & 0.0 & 100.0 & 0.0 & 0.0 & 100.0 \\
10 & 0.0 & 0.0 & 100.0 & 0.0 & 0.0 & 100.0 \\
20 & 0.0 & 0.0 & 100.0 & 0.0 & 0.4 & 99.6 \\
50 & 0.0 & 0.0 & 100.0 & 0.0 & 36.8 & 63.2 \\
100 & 0.0 & 0.0 & 100.0 & 0.0 & 36.8 & 63.2 \\
200 & 0.0 & 0.0 & 100.0 & 0.0 & 39.0 & 61.0 \\
500 & 0.0 & 0.0 & 100.0 & 0.0 & 63.2 & 36.8 \\
1,000 & 0.0 & 0.0 & 100.0 & 0.0 & 63.2 & 36.8 \\
2,500 & 62.5 & 0.0 & 37.5 & 0.0 & 63.2 & 36.8 \\
5,000 & 100.0 & 0.0 & 0.0 & 0.0 & 100.0 & 0.0 \\
10,000 & 100.0 & 0.0 & 0.0 & 0.0 & 100.0 & 0.0 \\
20,000 & 100.0 & 0.0 & 0.0 & 0.0 & 100.0 & 0.0 \\
\bottomrule
\end{tabular}
\end{table}

\begin{table}[t]
\centering
\scriptsize
\caption{Controlled-grid Hoeffding monitor status rates at every evaluated look.}
\label{tab:controlled-status-rates}
\setlength{\tabcolsep}{3.4pt}
\begin{tabular}{r rrr rrr}
\toprule
& \multicolumn{3}{c}{Safe world (\%)} & \multicolumn{3}{c}{Harmful world (\%)} \\
Look & Deploy & Hold & Cannot & Deploy & Hold & Cannot \\
\midrule
5 & 0.0 & 0.0 & 100.0 & 0.0 & 0.0 & 100.0 \\
10 & 0.0 & 0.0 & 100.0 & 0.0 & 0.0 & 100.0 \\
20 & 0.0 & 0.0 & 100.0 & 0.0 & 0.0 & 100.0 \\
50 & 0.0 & 0.0 & 100.0 & 0.0 & 0.0 & 100.0 \\
100 & 0.0 & 0.0 & 100.0 & 0.0 & 0.0 & 100.0 \\
200 & 0.0 & 0.0 & 100.0 & 0.0 & 0.0 & 100.0 \\
500 & 0.0 & 0.0 & 100.0 & 0.0 & 0.0 & 100.0 \\
1,000 & 0.0 & 0.0 & 100.0 & 0.0 & 0.4 & 99.6 \\
2,500 & 5.3 & 0.0 & 94.7 & 0.0 & 5.9 & 94.1 \\
5,000 & 11.0 & 0.0 & 89.0 & 0.0 & 11.0 & 89.0 \\
10,000 & 19.0 & 0.0 & 81.0 & 0.0 & 19.3 & 80.7 \\
20,000 & 30.0 & 0.0 & 70.0 & 0.0 & 30.1 & 69.9 \\
\bottomrule
\end{tabular}
\end{table}

\begin{table}[t]
\centering
\scriptsize
\caption{Fixed-panel first-decision stopping-time distribution, including right-censoring after 20,000 observations.}
\label{tab:fixed-stopping-rates}
\setlength{\tabcolsep}{4pt}
\begin{tabular}{r rr rr}
\toprule
& \multicolumn{2}{c}{Safe world (\%)} & \multicolumn{2}{c}{Harmful world (\%)} \\
First decision & Deploy & Hold & Deploy & Hold \\
\midrule
5 & 0.0 & 0.0 & 0.0 & 0.0 \\
10 & 0.0 & 0.0 & 0.0 & 0.0 \\
20 & 0.0 & 0.0 & 0.0 & 0.4 \\
50 & 0.0 & 0.0 & 0.0 & 36.4 \\
100 & 0.0 & 0.0 & 0.0 & 0.0 \\
200 & 0.0 & 0.0 & 0.0 & 2.2 \\
500 & 0.0 & 0.0 & 0.0 & 24.3 \\
1,000 & 0.0 & 0.0 & 0.0 & 0.0 \\
2,500 & 62.5 & 0.0 & 0.0 & 0.0 \\
5,000 & 37.5 & 0.0 & 0.0 & 36.8 \\
10,000 & 0.0 & 0.0 & 0.0 & 0.0 \\
20,000 & 0.0 & 0.0 & 0.0 & 0.0 \\
Right-censored & \multicolumn{2}{c}{0.0} & \multicolumn{2}{c}{0.0} \\
\bottomrule
\end{tabular}
\end{table}

\begin{table}[t]
\centering
\scriptsize
\caption{Controlled-grid first-decision stopping-time distribution, including right-censoring after 20,000 observations.}
\label{tab:controlled-stopping-rates}
\setlength{\tabcolsep}{4pt}
\begin{tabular}{r rr rr}
\toprule
& \multicolumn{2}{c}{Safe world (\%)} & \multicolumn{2}{c}{Harmful world (\%)} \\
First decision & Deploy & Hold & Deploy & Hold \\
\midrule
5 & 0.0 & 0.0 & 0.0 & 0.0 \\
10 & 0.0 & 0.0 & 0.0 & 0.0 \\
20 & 0.0 & 0.0 & 0.0 & 0.0 \\
50 & 0.0 & 0.0 & 0.0 & 0.0 \\
100 & 0.0 & 0.0 & 0.0 & 0.0 \\
200 & 0.0 & 0.0 & 0.0 & 0.0 \\
500 & 0.0 & 0.0 & 0.0 & 0.0 \\
1,000 & 0.0 & 0.0 & 0.0 & 0.4 \\
2,500 & 5.3 & 0.0 & 0.0 & 5.5 \\
5,000 & 5.7 & 0.0 & 0.0 & 5.1 \\
10,000 & 8.0 & 0.0 & 0.0 & 8.4 \\
20,000 & 11.0 & 0.0 & 0.0 & 10.8 \\
Right-censored & \multicolumn{2}{c}{70.0} & \multicolumn{2}{c}{69.9} \\
\bottomrule
\end{tabular}
\end{table}

\begin{table}[t]
\centering
\scriptsize
\caption{Fixed-panel final-look status comparison at 20,000 attempted observations. Wilson is pointwise only; both confidence-sequence rows are stopping-valid under the declared model. Unsupported was 0.0\% throughout.}
\label{tab:method-comparison}
\setlength{\tabcolsep}{3.4pt}
\begin{tabular}{l rrr rrr}
\toprule
& \multicolumn{3}{c}{Safe world (\%)} & \multicolumn{3}{c}{Harmful world (\%)} \\
Method & Deploy & Hold & Cannot & Deploy & Hold & Cannot \\
\midrule
Hoeffding CS & 100.0 & 0.0 & 0.0 & 0.0 & 100.0 & 0.0 \\
Wilson pointwise & 100.0 & 0.0 & 0.0 & 0.0 & 100.0 & 0.0 \\
Beta-binomial CS & 100.0 & 0.0 & 0.0 & 0.0 & 100.0 & 0.0 \\
\bottomrule
\end{tabular}
\end{table}

\begin{table}[t]
\centering
\scriptsize
\caption{Controlled-grid decision-onset comparison. Counts in parentheses are numbers of the 18 prespecified cells whose first qualifying inspected look occurred at that value. Wilson onsets are descriptive because Wilson is pointwise.}
\label{tab:decision-onset-comparison}
\setlength{\tabcolsep}{4pt}
\begin{tabularx}{\textwidth}{l c c Y}
\toprule
Method & Stopping-valid & Qualified & First qualifying inspected looks \\
\midrule
Hoeffding CS & yes & 4/18 & 2,500 (1), 5,000 (1), 20,000 (2) \\
Wilson pointwise & no & 12/18 & 500 (1), 1,000 (1), 2,500 (3), 5,000 (1), 10,000 (4), 20,000 (2) \\
Beta-binomial CS & yes & 9/18 & 1,000 (1), 2,500 (1), 5,000 (2), 10,000 (2), 20,000 (3) \\
\bottomrule
\end{tabularx}
\end{table}

Complete method-by-cell status and timing rows for all three methods are released in machine-readable form.

%% file: checklist.tex
\section*{NeurIPS Paper Checklist}

\begin{enumerate}

\item {\bf Claims}
    \item[] Question: Do the main claims made in the abstract and introduction accurately reflect the paper's contributions and scope?
    \item[] Answer: \answerYes{}
    \item[] Justification: The abstract, introduction, and limitations state the matched-world result, conditional monitoring result, fixed-panel scope, and controlled-panel power limitation.

\item {\bf Limitations}
    \item[] Question: Does the paper discuss the limitations of the work performed by the authors?
    \item[] Answer: \answerYes{}
    \item[] Justification: The Scope paragraph states that the benchmark is synthetic, makes no production-prevalence claim, treats the fixed panel descriptively, does not equate zero observed false-safe cases with zero risk, and conditions monitoring conclusions on the declared sampling model.

\item {\bf Theory assumptions and proofs}
    \item[] Question: For each theoretical result, does the paper provide the full set of assumptions and a complete (and correct) proof?
    \item[] Answer: \answerYes{}
    \item[] Justification: Theorem~\ref{thm:equivalence} is explicitly positioned as an application of standard identification and two-point indistinguishability ideas, states all assumptions, gives a proof sketch, and Appendix~\ref{app:proof} provides the complete proof.

    \item {\bf Experimental result reproducibility}
    \item[] Question: Does the paper fully disclose all the information needed to reproduce the main experimental results of the paper to the extent that it affects the main claims and/or conclusions of the paper (regardless of whether the code and data are provided or not)?
    \item[] Answer: \answerYes{}
    \item[] Justification: The paper and public source release report the scoring rules, statistical units, complete generator and experiment configurations, seeds, stream counts, inspected looks, thresholds, assumptions, decision criteria, exact commands, tests, and released result records.

\item {\bf Open access to data and code}
    \item[] Question: Does the paper provide open access to the data and code, with sufficient instructions to faithfully reproduce the main experimental results, as described in supplemental material?
    \item[] Answer: \answerYes{}
    \item[] Justification: The public code release contains the executable source, data, configurations, exact commands, tests, figures, MIT project and dataset license, and retained third-party notice needed to reproduce the main results.

\item {\bf Experimental setting/details}
    \item[] Question: Does the paper specify all the training and test details (e.g., data splits, hyperparameters, how they were chosen, type of optimizer) necessary to understand the results?
    \item[] Answer: \answerYes{}
    \item[] Justification: The paper and released configurations give every generator parameter, seed, cohort size, nested-stream aggregation rule, monitoring look, threshold, interval method, and qualification rule; there is no training or optimization.

\item {\bf Experiment statistical significance}
    \item[] Question: Does the paper report error bars suitably and correctly defined or other appropriate information about the statistical significance of the experiments?
    \item[] Answer: \answerYes{}
    \item[] Justification: Fixed heterogeneous panels are explicitly descriptive; the comparisons distinguish pointwise Wilson fixed-sample intervals from anytime-valid sequential bounds, and Section 4 and Appendix~\ref{app:experiments} state that first-look selection is not a formal sample-complexity estimate or simultaneous pointwise inference.

\item {\bf Experiments compute resources}
    \item[] Question: For each experiment, does the paper provide sufficient information on the computer resources (type of compute workers, memory, time of execution) needed to reproduce the experiments?
    \item[] Answer: \answerYes{}
    \item[] Justification: Appendix~\ref{app:experiments} reports the CPU model and architecture, installed memory, Python version, worker count, wall and CPU time, maximum resident set, output size, and absence of GPU, network, or model-service use for the central released command; it also states that unrecorded development and failed exploratory CPU runs are excluded.

\item {\bf Code of ethics}
    \item[] Question: Does the research conducted in the paper conform, in every respect, with the NeurIPS Code of Ethics \url{https://neurips.cc/public/EthicsGuidelines}?
    \item[] Answer: \answerYes{}
    \item[] Justification: The work uses only synthetic environments, contains no human or personal data, and follows the NeurIPS Code of Ethics.

\item {\bf Broader impacts}
    \item[] Question: Does the paper discuss both potential positive societal impacts and negative societal impacts of the work performed?
    \item[] Answer: \answerYes{}
    \item[] Justification: Appendix~\ref{app:experiments} discusses reduced unsupported deployment, delayed beneficial updates, omitted groups, and appropriate human governance.

\item {\bf Safeguards}
    \item[] Question: Does the paper describe safeguards that have been put in place for responsible release of data or models that have a high risk for misuse (e.g., pre-trained language models, image generators, or scraped datasets)?
    \item[] Answer: \answerNA{}
    \item[] Justification: The release contains small synthetic records and source code, not a trained model, scraped dataset, personal data, or high-risk dual-use asset.

\item {\bf Licenses for existing assets}
    \item[] Question: Are the creators or original owners of assets (e.g., code, data, models), used in the paper, properly credited and are the license and terms of use explicitly mentioned and properly respected?
    \item[] Answer: \answerYes{}
    \item[] Justification: The paper cites every diagnostic method; the only incorporated third-party implementation is MIT-licensed SeqE-Guard logic, and its license notice is retained with the source.

\item {\bf New assets}
    \item[] Question: Are new assets introduced in the paper well documented and is the documentation provided alongside the assets?
    \item[] Answer: \answerYes{}
    \item[] Justification: The manuscript and MIT-licensed public release document generation, scoring, assumptions, workload sizes, and statistical interpretation for the new synthetic benchmark.

\item {\bf Crowdsourcing and research with human subjects}
    \item[] Question: For crowdsourcing experiments and research with human subjects, does the paper include the full text of instructions given to participants and screenshots, if applicable, as well as details about compensation (if any)?
    \item[] Answer: \answerNA{}
    \item[] Justification: The work contains no crowdsourcing or research with human subjects.

\item {\bf Institutional review board (IRB) approvals or equivalent for research with human subjects}
    \item[] Question: Does the paper describe potential risks incurred by study participants, whether such risks were disclosed to the subjects, and whether Institutional Review Board (IRB) approvals (or an equivalent approval/review based on the requirements of your country or institution) were obtained?
    \item[] Answer: \answerNA{}
    \item[] Justification: The work contains no crowdsourcing or research with human subjects, so institutional review was not applicable.

\item {\bf Declaration of LLM usage}
    \item[] Question: Does the paper describe the usage of LLMs if it is an important, original, or non-standard component of the core methods in this research? Note that if the LLM is used only for writing, editing, or formatting purposes and does \emph{not} impact the core methodology, scientific rigor, or originality of the research, declaration is not required.
    \item[] Answer: \answerNA{}
    \item[] Justification: Language models are not an original or non-standard component of the benchmark, exact evaluators, monitor, or experiments; any writing assistance does not affect the methodology, data, scoring, or reported results.

\end{enumerate}